\documentclass[11pt,a4paper]{article}

\usepackage[utf8]{inputenc}
\usepackage[T1]{fontenc}
\usepackage{amsmath,amssymb,amsthm}
\usepackage{graphicx}
\usepackage[colorlinks=true,linkcolor=blue,citecolor=blue,urlcolor=blue]{hyperref}
\usepackage[numbers]{natbib}
\usepackage{geometry}
\usepackage{booktabs}
\usepackage{algorithm}
\usepackage{algorithmic}
\usepackage{xcolor}
\usepackage{float}

\newtheorem{definition}{Definition}
\newtheorem{proposition}{Proposition}
\newtheorem{remark}{Remark}
\newtheorem{principle}{Principle}

\title{\textbf{NRR-Phi: A Typed External Text-to-State Interface and Update Contract for Inspectable Ambiguity-State Maintenance}}

\author{
  Kei Saito\thanks{ORCID: 0009-0006-4715-9176} \\
  Independent Researcher, Japan \\
  \texttt{kei.saito.research@gmail.com}
}

\date{First posted 12 January 2026; revised July 2026 \\[0.5em]
\textit{Part of the Non-Resolution Reasoning (NRR) research program.}}

\begin{document}

\maketitle

\begin{center}
\textcopyright\ 2026 Kei Saito.
Licensed under CC BY 4.0.\\
\url{https://creativecommons.org/licenses/by/4.0/}
\end{center}

\begin{abstract}
Ambiguity-bearing inputs often reach downstream systems through interfaces that favor a
single resolved response before later context arrives. Even when candidate alternatives
are externalized, their continued representation and relative activation across later
updates depend on the update rule. We
address this state-maintenance problem within Non-Resolution Reasoning (NRR) by specifying
a typed external text-to-state interface together with an explicit state-update contract.

We present a formal framework for \textit{text-to-state mapping} ($\phi: \mathcal{T} \to \mathcal{S}$)
that maps extracted items into typed $(v,c,w,m)$ records. We make operator-defined
record retention inspectable and define a record-weight entropy criterion that detects
concentration in the normalized record weights. The mapping decomposes into conflict
detection, retained-item extraction, and
state construction; the update layer specifies dampening, calibration, deferred
resolution, contradiction-preserving integration, and temporal persistence.

Across 580 constructed states/pairs, an executable suite performs 2,740
operator--state measurements. The tested non-violating transitions and
calibration/identity checks exhibit 0\% record-weight
entropy violations, while a uniform-subtraction comparison violates the criterion at
rates that rise from 1.7\% to 6.1\% and 17.8\% as subtraction increases. A separate
68-input construction audit shows that the reported rule-based and archived
LLM-assisted procedures populate the typed schema with multiple positive-weight records
(mean record-weight entropy $H = 1.087$ bits), including a separate Japanese marker-set
instantiation. The operator result is executable contract-conformance evidence with a
discriminating negative control. The entropy criterion does not by itself certify
record identity or cardinality, semantic adequacy, or end-to-end behavioral improvement.

\textbf{Package surface}: The current repository bundles deterministic reruns for the
rule-based mapping and operator-validation appendix, together with archived prompt/output
artifacts, a transcript-audit path, and a fixed 18-set sanity-check rerun summary for
the LLM-handled ambiguity cases reported here.

Together, Phi turns retained-state maintenance into a typed, executable, and
falsifiable interface: record carry-forward is inspectable in declared operators,
while normalized-weight concentration is separately testable before a downstream
policy commits.
\end{abstract}


\section{Introduction}
\label{sec:intro}

This paper develops the text-to-state mapping layer of the Non-Resolution Reasoning (NRR) framework. It focuses on how heterogeneous extracted items are mapped into a retained state, and it should be read alongside companion papers that address foundational criteria, implementation, operational calibration, and bounded verification.

\subsection{The Early Semantic Commit Problem}

Large language models (LLMs) have achieved remarkable capabilities in natural language 
understanding and generation. However, deployed inference interfaces typically return
one resolved response, which creates pressure toward early commitment when context is
still incomplete.

Consider the input: ``I want to quit my job, but I also don't want to quit.'' 
A standard one-response interface must produce an output, which pushes the system toward
one pole of this ambivalent statement. Typical responses either:
\begin{itemize}
    \item Select one pole of the ambivalence as primary (``It sounds like you're leaning toward staying...'')
    \item Reframe the ambivalence as a problem requiring resolution (``Let's make a pros and cons list...'')
    \item Request clarification (``Which feeling is stronger?'')
\end{itemize}

Each of these responses treats the ambivalence as something to resolve. They may
enumerate alternatives in text, but they do not by themselves maintain both
attitude records as one persistent state for later operators.

This paper does not attribute early commitment to any single internal mechanism.
Autoregressive interfaces emit one token sequence, but neither attention softmax,
autoregressive decoding, nor cross-entropy training alone establishes that a model has
discarded alternative interpretations internally. Our claim is at the system-interface
level: unless extracted alternatives are externalized into a persistent structure, a
generated response does not by itself provide a typed, operator-addressable record of
them for later updates.

\subsection{Non-Resolution Reasoning}

Non-Resolution Reasoning (NRR) \cite{saito2025nrr} treats ambiguity retention as a valid
computational mode. Core proposed the state space
$\mathcal{S} = \{(v_i, c_i, w_i, m_i)\}$ and reported a controlled synthetic
gated multi-vector experiment: NRR-lite produced Turn-1 output entropy
$H = 0.91$ bits, while the tested single-embedding baseline produced
$H = 0.15$ bits, with both tested systems resolving correctly after context arrived.

However, two connected engineering questions remained open within NRR: \textbf{how does
text become typed retained state, and which update rules carry candidate records forward
without concentrating their relative activation?} Phi specifies both the mapping
$\phi: \mathcal{T} \to \mathcal{S}$ and the operator contract applied after construction.

\subsection{Contributions}

This paper specifies the previously open text-to-state and state-maintenance interface
within the NRR framework. Our contributions are:

\begin{enumerate}
    \item \textbf{Typed Text-to-State Interface}: We define $\phi$ as conflict
          detection, retained-item extraction, and state construction into typed
          $(v,c,w,m)$ records, with rule-based and archived LLM-assisted item sources.

    \item \textbf{Inspectable Update Contract}: We define a record-weight entropy
          preservation criterion and a declared family of state transitions whose
          effects on retained records can be proved or tested.

    \item \textbf{Executable Operator Conformance Suite}: Across 580 constructed states/pairs
          and 2,740 measurements, the declared transitions and checks are executed against
          the entropy criterion: tested non-violating rows show
          0\% criterion violations, while uniform subtraction rises from 1.7\% to 17.8\%
          as subtraction magnitude increases.

    \item \textbf{Construction Support}: A mixed 68-input audit shows that the reported
          extraction procedures populate multi-record states (mean $H = 1.087$ bits),
          including a separate Japanese marker-set instantiation. This supports schema
          construction rather than semantic adequacy.
\end{enumerate}

The positive center is an engineering contract: once candidate items are externalized,
Phi makes operator-defined record retention and normalized weight concentration
separately inspectable.

\paragraph{Support structure.}
The evidence surface has three roles. The typed mapping and operator definitions state
the contract, including which records each declared operator carries forward. The
580-unit operator audit and its negative control test conformance to the declared
record-weight invariant. The 68-input audit, archived transcript path,
and fixed 18-set sanity-check rerun show that the reported extraction procedures
populate the schema. Record-weight conformance is structural and does not adjudicate
whether the extracted candidates are semantically correct, distinct, or exhaustive.

\subsection{Paper Organization}

Section~\ref{sec:related} reviews related work on uncertainty in LLM inference. 
Section~\ref{sec:framework} presents the formal framework for text-to-state mapping. 
Section~\ref{sec:algorithm} provides algorithms and implementation details. 
Section~\ref{sec:experiments} reports the state-construction audit.
Section~\ref{sec:operator-validation} presents the operator-contract validation and its
main result.
Section~\ref{sec:application} briefly illustrates application to psychological support contexts. 
Section~\ref{sec:discussion} discusses limitations and future work. 
Section~\ref{sec:conclusion} concludes. 
Appendix~\ref{app:operators} gives the full operator definitions, proofs, validation
method, figures, and extended results.

\paragraph{Scope and series alignment.}
In this paper, $\sigma_{\mathrm{state}}$ denotes a state-level calibration operator in Appendix~\ref{app:operators}, distinct from the item-level strengthen/dampen operators ($\sigma_{\text{item}}, \delta_{\text{item}}$) used in later implementation papers. Phi evaluates state construction and structural non-collapse properties, especially record-weight entropy and operator violation rates; later implementation papers add interface and operational metrics such as extraction reliability and token efficiency. These metrics are complementary layers rather than competing definitions of success. The present claim is limited to the typed external state and tested operator contract, not semantic correctness or downstream behavioral improvement.

\section{Related Work}
\label{sec:related}

\subsection{Uncertainty in LLM Inference}

Uncertainty quantification in neural networks has received substantial attention 
\cite{gal2016dropout, lakshminarayanan2017simple}. For LLMs specifically, calibration 
studies examine whether model confidence aligns with accuracy \cite{kadavath2022language}. 
These approaches measure uncertainty in model predictions or parameters. Phi instead
specifies an external record structure for items supplied by an extraction procedure.

Epistemic uncertainty (model uncertainty) and aleatoric uncertainty (data uncertainty) 
are typically distinguished \cite{hullermeier2021aleatoric}. Phi addresses a different
system-interface dimension: whether an extraction procedure returns multiple state items
and whether those items remain addressable as records. Their validity requires separate evaluation.

\subsection{Ambiguity Recognition, Clarification, and Revision}

Substantial prior work already treats ambiguity as an object to represent, detect, enumerate, or resolve. Packed machine-translation representations and underspecified semantic representations preserve compact sets of analyses without requiring immediate enumeration of every resolved reading \cite{emele1998packed,alshawi2011underspecified,manshadi2018coherence}. Information-state dialogue management explicitly combines an information state, update rules, and dialogue control \cite{larsson2000information}; POMDP dialogue systems maintain uncertain state hypotheses and select actions through a policy \cite{williams2007pomdp}. AmbigQA requires every plausible answer and a disambiguated question for each answer \cite{min2020ambigqa}. CLAM supplies a selective-clarification pipeline for ambiguous questions \cite{kuhn2023clam}; AmbiEnt tests recognition and disentanglement of sentence meanings \cite{liu2023ambient}; DUST tests detection and interpretation of semantic underspecification and finds little model uncertainty during interpretation \cite{wildenburg2024dust}; and Alignment with Perceived Ambiguity trains models to detect and manage ambiguity while retaining clear-question performance \cite{kim2024apa}. Clarification work estimates when a question is useful from model or intent uncertainty \cite{testoni2024asking,zhang2025clarify}, while a semantic-parsing pipeline generates missing natural-language interpretations before mapping them to logical forms \cite{saparina2025disambiguate}. These studies establish that ambiguity representation, explicit state, update/control architectures, detection, enumeration, and management predate $\phi$.

Phi's narrower distinction is an NRR-specific heterogeneous external record schema and operator contract: extracted items are typed as $(v,c,w,m)$ records, may remain independently addressable, and can be projected to an output without the projection definition deleting the retained collection. The present paper does not claim the first explicit language or dialogue state, the first parallel hypothesis set, or the first state-update/action-selection architecture, and it does not establish comparative superiority over those predecessors.

Adjacent work addresses temporal update. Belief-R tests whether later premises trigger appropriate revision without inducing unnecessary change \cite{wilie2024belief}. An arXiv preprint first posted in May 2025, and later published as an ICLR 2026 Outstanding Paper, finds that models make early assumptions, attempt solutions prematurely, and often fail to recover \cite{laban2026lost}. In referentially ambiguous dialogue, models frequently commit to one reading or cover all candidates rather than hedge or clarify, with simplified-language instructions further reducing diverse response strategies \cite{ellinger2025depends}.

After Phi v1 (12 January 2026), work appearing in this problem family named \emph{ambiguity collapse} as an epistemic risk \cite{gurarieh2026collapse}, measured premature closure in clinical decisions \cite{handler2026closure}, tracked commitment and recovery in multi-turn diagnosis \cite{fang2026mint}, studied premature confidence within reasoning trajectories \cite{gai2026confidence}, and proposed hidden-state diagnostics of premature commitment in agents \cite{mehta2026commitment}. Phi addresses a complementary engineering layer: extracted items become typed $(v,c,w,m)$ records, and declared update rules are checked for suppressing those records under an explicit invariant.

\subsection{Multi-Hypothesis Approaches}

Beam search and diverse decoding \cite{vijayakumar2016diverse} 
maintain multiple candidates during generation but ultimately select a single output. 
Mixture-of-Experts architectures \cite{shazeer2017outrageously, fedus2021switch} route
computation through specialized subnetworks while the system still realizes an output.

These approaches provide candidate or computational multiplicity for different objectives. Phi's distinction is an external typed-record contract whose update rules expose both record-collection behavior and normalized weight concentration, while a singular output can be projected without deleting the external state by definition.

\subsection{Ambiguity in NLP}

Word sense disambiguation \cite{navigli2009word}, coreference resolution \cite{lee2017end}, and pragmatic inference \cite{goodman2016pragmatic} address distinct resolution problems. They are not treated here as failed versions of Phi; Phi specifies a different system-level state interface for extracted items that remain live.

\subsection{Relation to NRR Framework}

This paper builds directly on NRR-Core \cite{saito2025nrr}. That work proposed the
state-space formalism and reported a synthetic gated multi-vector proof-of-concept
(NRR-lite output entropy $H = 0.91$ bits vs.\ $H = 0.15$ bits for a controlled
single-embedding baseline under the tested setup).
NRR defines a transition-oriented core operator set ($\delta, \tau, \kappa, \pi$), with
item-level strengthening available in later implementations, and additional
metadata functions ($\alpha, \rho, \iota$) for representation management
(see Appendix~\ref{app:operators} for summary).

This paper specifies the NRR-specific text-to-state interface: without a defined $\phi$,
the proposed NRR operators lack a declared input transformation.

\section{Formal Framework}
\label{sec:framework}

\subsection{Preliminary Definitions}

\begin{definition}[Text Space]
Let $\mathcal{T}$ denote the space of natural language texts, where each 
$T \in \mathcal{T}$ is a sequence of tokens $(t_1, t_2, \ldots, t_n)$.
\end{definition}

\begin{definition}[Retained State Space]
Following \cite{saito2025nrr}, a \textit{retained state space} is a structure 
$\mathcal{S} = (V, C, W, M, \preceq)$ where:
\begin{itemize}
    \item $V \subseteq \mathbb{R}^d$ is a set of semantic vector representations
    \item $C$ is a set of context identifiers
    \item $W: V \times C \to \mathbb{R}_{\geq 0}$ is a nonnegative activation weight function
    \item $M$ is a set of metadata flags (including conflict markers)
    \item $\preceq$ is a partial order on $V$ (subsumption relation)
\end{itemize}
A state instance is a set $S = \{(v_i, c_i, w_i, m_i)\}_{i=1}^{n}$ drawn from this space.
It is an external record collection: each item is tagged by context, weight, and metadata and remains available to declared operators until an operation removes or changes it.
This is the same nonnegative, not-necessarily-bounded weight contract used by the
revised Core specification; bounded scores are permitted as an implementation subset.
\end{definition}

\begin{definition}[Retained Item and Record]
A \textit{retained item} $x$ may be a proposed semantic reading, a proposition or
attitude, an epistemic alternative, or a marker-delimited segment. Its state record is a
tuple $r=(v,c,w,m)$ where:
\begin{itemize}
    \item $v \in V$ encodes the item content
    \item $c \in C$ identifies the context under which the item was produced
    \item $w \in \mathbb{R}_{\geq 0}$ is its activation weight
    \item $m \in M$ contains source, conflict, and item-type metadata
\end{itemize}
An \textit{interpretation} is the semantic-reading subtype: an item that proposes an
alternate reading of an expression. Let $\mathcal{R}(T)$ index all retained items
constructed from $T$, and let $\mathcal{I}(T)\subseteq\mathcal{R}(T)$ index those typed as
candidate semantic readings. Segments created by adversative or hedging markers enter
$\mathcal{R}(T)$ as segment, proposition, or attitude records unless separate adjudication
supports the semantic-reading type. Marker presence is not proof that the input has
multiple meanings.
\end{definition}

\subsection{The Text-to-State Mapping}

\begin{definition}[Text-to-State Mapping]
The mapping $\phi: \mathcal{T} \to \mathcal{S}$ transforms input text $T$ 
to a state space instance:
\begin{equation}
\phi(T) = \{(v_i, c_i, w_i, m_i) \mid i \in \mathcal{R}(T)\}
\end{equation}
where $\mathcal{R}(T)$ is the retained-item index set constructed from $T$.
\end{definition}

The mapping $\phi$ decomposes into three sequential operations:
\begin{equation}
\phi = \psi_{\text{state}} \circ \psi_{\text{item}} \circ \psi_{\text{conflict}}
\end{equation}
where:
\begin{enumerate}
    \item $\psi_{\text{conflict}}: \mathcal{T} \to \mathcal{T} \times \{0,1\}^k$ 
          detects conflict markers
    \item $\psi_{\text{item}}: \mathcal{T} \times \{0,1\}^k \to \mathcal{R}_{\mathrm{typed}}$ 
          constructs typed retained items
    \item $\psi_{\text{state}}: \mathcal{R}_{\mathrm{typed}} \to \mathcal{S}$ 
          constructs the state space
\end{enumerate}

\subsection{Stage 1: Conflict Detection}

The conflict detection stage identifies linguistic markers that can license
segmentation or candidate generation; it does not by itself establish multiple meanings.

\begin{definition}[Conflict Markers]
A \textit{conflict marker set} $\mathcal{M}_{\text{conf}}$ is a collection of
linguistic patterns used to trigger segmentation or candidate generation:
\begin{equation}
\mathcal{M}_{\text{conf}} = \mathcal{M}_{\text{explicit}} \cup \mathcal{M}_{\text{implicit}} \cup \mathcal{M}_{\text{structural}}
\end{equation}
where:
\begin{itemize}
    \item $\mathcal{M}_{\text{explicit}}$: Explicit contradiction markers 
          (e.g., ``but'', ``however'', ``on the other hand'', ``yet'')
    \item $\mathcal{M}_{\text{implicit}}$: Hedging and uncertainty markers 
          (e.g., ``maybe'', ``perhaps'', ``might'', ``I think'')
    \item $\mathcal{M}_{\text{structural}}$: Structural ambiguity indicators 
          (e.g., ``both X and Y'', ``either... or'', coordination ambiguity)
\end{itemize}
The marker set is separately re-instantiated for the Japanese cases in this paper;
corresponding markers include adversatives (\textit{kedo}, \textit{demo},
\textit{shikashi}), hedging expressions (\textit{kamoshirenai}), and epistemic markers
(\textit{to omou}). See Appendix for the complete taxonomy.
\end{definition}

\begin{definition}[Conflict Detection Function]
The conflict detection function $\psi_{\text{conflict}}$ returns:
\begin{equation}
\psi_{\text{conflict}}(T) = (T, \mathbf{f})
\end{equation}
where $\mathbf{f} = (f_1, f_2, \ldots, f_k) \in \{0,1\}^k$ is a binary feature 
vector indicating presence of each marker type.
\end{definition}

\subsection{Stage 2: Retained-Item Extraction}

Retained-item extraction transforms text into a typed collection using a hybrid approach.
Rule-based segmentation produces marker-delimited segment, proposition, or attitude
items; LLM-assisted enumeration produces candidate semantic readings or epistemic
alternatives. These source labels are recorded in metadata and are not semantic
adequacy judgments.

\begin{definition}[Hybrid Extraction]
The retained-item extraction function combines two typed mechanisms:
\begin{equation}
\begin{split}
\psi_{\text{rule}}(T,\mathbf{f})
  &= \text{TypeSegments}(\text{SegmentAtMarkers}(T,\mathbf{f}),\mathbf{f}),\\
\psi_{\text{LLM}}(T,\mathbf{f})
  &= \text{TypeCandidates}(\text{LLM}_{\theta}(\text{Prompt}_{\text{interp}}(T,\mathbf{f})),\mathbf{f}),\\
\psi_{\text{item}}(T, \mathbf{f})
  &= \text{Merge}(\psi_{\text{rule}}(T, \mathbf{f}), \psi_{\text{LLM}}(T, \mathbf{f})).
\end{split}
\end{equation}
Both typed branches return records in
$\mathcal{R}_{\mathrm{typed}}=\{(x,\mathrm{ctx},\mathrm{conf},\mathrm{type})\}$,
with $\mathrm{conf}\in\mathbb{R}_{\geq0}$ and a declared item type.
\end{definition}

\paragraph{Rule-Based Extraction.}
For texts with detected conflict markers ($\|\mathbf{f}\|_1 > 0$), 
rule-based extraction segments text at conflict boundaries:
\begin{equation}
\text{SegmentAtMarkers}(T, \mathbf{f}) = \{(s_j, c_j) \mid j \in J(T, \mathbf{f})\}
\end{equation}
where $J(T, \mathbf{f})$ indexes segments delimited by conflict markers,
$s_j$ is the $j$-th segment, and $c_j$ is the context label derived from
marker type (e.g., ``pre-adversative'', ``post-adversative'', ``hedge-scope'').
The deterministic typing function then assigns
\begin{equation}
\text{TypeSegments}(\{(s_j,c_j)\},\mathbf{f})
=\{(s_j,c_j,q_{\mathrm{rule}}(s_j,c_j,\mathbf{f}),
t_{\mathrm{rule}}(c_j))\},
\end{equation}
where the declared rule score $q_{\mathrm{rule}}\geq0$ supplies the activation
input and the fixed marker-to-type lookup
$t_{\mathrm{rule}}$ returns \texttt{segment}, \texttt{proposition}, or
\texttt{attitude}. The rule configuration fixes $q_{\mathrm{rule}}$ before state
construction (as a constant or marker-specific score) and records that choice;
merge does not infer it. These source types do not assert semantic adequacy.

\paragraph{LLM-Based Extraction.}
For complex or implicit ambiguity, LLM-based extraction prompts a language 
model to propose candidate readings or epistemic alternatives:
The structured LLM response is parsed as raw tuples
$(x_k,c_k,q_k)$, where $q_k\geq0$ is the returned confidence field.
$\text{TypeCandidates}$ applies a fixed route-to-type lookup: a requested lexical
or structural reading is tagged \texttt{candidate-}\allowbreak\texttt{reading}, and a requested
epistemic alternative is tagged \texttt{epistemic-}\allowbreak\texttt{alternative}. It returns
$(x_k,c_k,q_k,t_{\mathrm{LLM}}(T,\mathbf{f}))$.
$\text{Prompt}_{\text{interp}}$ requests the raw fields needed by this typed path.
``ALL'' in the archived prompt is an elicitation instruction, not evidence that
the returned set is exhaustive, and neither $q_k$ nor $q_{\mathrm{rule}}$ is
claimed to be calibrated plausibility.

\paragraph{Merge Operation.}
The merge function removes duplicates via semantic similarity:
\begin{equation}
\begin{aligned}
\operatorname{Merge}(R_1,R_2)
  &= R_1 \cup \{r \in R_2 \mid {} \\
  &\qquad \forall q \in R_1:\operatorname{sim}(r,q)<\tau\}.
\end{aligned}
\end{equation}
where $\text{sim}(r, q)$ is cosine similarity in a sentence-level embedding space
(for example, SentenceBERT \cite{reimers2019sentencebert}),
and $\tau \in (0,1)$ is a conceptual deduplication threshold describing the
merge contract. The current paper does not treat one fixed executed $\tau$
value as a separately validated package claim.

\subsection{Stage 3: State Construction}

\begin{definition}[State Construction]
Given retained items $\mathcal{R} = \{(x_i, \text{ctx}_i, \text{conf}_i, \text{type}_i)\}$,
the state construction function produces:
\begin{equation}
\psi_{\text{state}}(\mathcal{R}) = \{(v_i, c_i, w_i, m_i)\}_{i=1}^{|\mathcal{R}|}
\end{equation}
where:
\begin{align}
v_i &= \text{Embed}(x_i) \in \mathbb{R}^d \\
c_i &= \text{ContextEncode}(\text{ctx}_i) \in C \\
w_i &= \text{conf}_i \cdot (1 + \beta \cdot \mathbf{1}[\text{conflict}]) \\
m_i &= \{\text{source}: \text{rule/LLM}, \text{type}: \text{type}_i,
        \text{conflict}: \mathbf{f}\}
\end{align}
\end{definition}

The factor $\beta > 0$ schematically increases activation for
records from conflicting contexts, reflecting the NRR principle
that conflict indicates information density. In the present paper, $\beta$
specifies the construction form rather than a separately audited executable
package constant.

\subsection{Information-Theoretic Properties}

\begin{definition}[State Entropy]
For a state $S = \{(v_i, c_i, w_i, m_i)\}_{i=1}^{n}$ with
$\sum_j w_j>0$, the \textit{state entropy} is the Shannon entropy of the
normalized weight distribution:
\begin{equation}
H(S) = -\sum_{i=1}^{n} p_i \log_2 p_i, \quad \text{where } p_i = \frac{w_i}{\sum_j w_j}
\end{equation}
Normalization is applied solely for entropy computation; the weights $w_i$ 
themselves remain unnormalized in the state representation.
The all-zero state is outside this entropy-evaluation domain.
\end{definition}

\begin{proposition}[Conditional Multi-Record Construction]
\label{thm:noncollapse}
If extraction and merge return at least two retained records with positive weights,
then state construction yields $|S| \geq 2$ and $H(S)>0$.
\end{proposition}

\begin{proof}
State construction creates one record for each returned item, so two or more returned
items give $|S|\geq2$. Positive weights normalize to a distribution with support on at
least two records, and therefore its Shannon entropy is strictly positive.
\end{proof}

\begin{remark}
This is a structural consequence of the construction, not a semantic-adequacy result.
It does not establish that the returned records are correct, exhaustive, mutually
distinct interpretations.
The $H=0$ single-record value used later is a stipulated reference point, not an
evaluated model baseline.
\end{remark}

\section{Algorithmic Form and Package Surface}
\label{sec:algorithm}

\subsection{Text-to-State Mapping Algorithm}

\begin{algorithm}[h]
\caption{Conceptual Text-to-State Mapping ($\phi$)}
\label{alg:phi}
\begin{algorithmic}[1]
\REQUIRE Text $T$, embedding function Embed, optional auxiliary candidate source
\ENSURE State $S = \{(v_i, c_i, w_i, m_i)\}$

\STATE \textbf{// Stage 1: Conflict Detection}
\STATE $\mathbf{f} \gets \text{DetectConflictMarkers}(T)$
\STATE $\text{has\_conflict} \gets (\|\mathbf{f}\|_1 > 0)$

\STATE \textbf{// Stage 2: Retained-Item Extraction}
\STATE $R_{\text{rule}} \gets \emptyset$
\IF{$\text{has\_conflict}$}
    \STATE $R_{\text{rule}} \gets \text{TypeSegments}(\text{SegmentAtMarkers}(T, \mathbf{f}))$
\ENDIF
\STATE $R_{\text{aux}} \gets \text{TypeCandidates}(\text{EnumerateImplicitAmbiguity}(T, \mathbf{f}))$
\STATE $R \gets \text{MergeContract}(R_{\text{rule}}, R_{\text{aux}})$

\STATE \textbf{// Stage 3: State Construction}
\STATE $S \gets \emptyset$
\FOR{$(x, \text{ctx}, \text{conf}, \text{type}) \in R$}
    \STATE $v \gets \text{Embed}(x)$
    \STATE $c \gets \text{ContextEncode}(\text{ctx})$
    \STATE $w \gets \text{AssignWeight}(\text{conf}, \text{has\_conflict})$
    \STATE $m \gets \{\text{conflict}: \mathbf{f}, \text{source}: \text{GetSource}(x),
                       \text{type}: \text{type}\}$
    \STATE $S \gets S \cup \{(v, c, w, m)\}$
\ENDFOR

\RETURN $S$
\end{algorithmic}
\end{algorithm}

Algorithm~\ref{alg:phi} describes the hybrid mapping contract at the design level.
The current package exposes deterministic reruns for the explicit-marker rule-based
mapping and for Appendix~\ref{app:operators} operator validation. For the implicit
ambiguity categories in Section~\ref{sec:experiments}, the package exposes archived
prompt/output transcripts, a sentence-level manifest, and a local audit script that
reconstruct the reported summaries rather than a fully rerunnable live LLM harness.
The rule-based segmentation results and archived prompted-enumeration results are
evaluated on disjoint subsets. The paper does not validate one end-to-end execution of
the conceptual rule-plus-LLM merge for every item.

\subsection{NRR Processing Pipeline}

The mapping $\phi$ feeds into the NRR operator pipeline (Appendix~\ref{app:operators}):

\begin{algorithm}[h]
\caption{NRR Processing Pipeline}
\label{alg:pipeline}
\begin{algorithmic}[1]
\REQUIRE Text $T$, previous state $S_{\text{prev}}$ (or $\emptyset$)
\ENSURE Processed state $S'$, output $o$

\STATE $S_{\text{new}} \gets \phi(T)$
\STATE $S_0 \gets \rho(S_{\text{new}})$ \COMMENT{Optional metadata tagging (weights unchanged)}
\STATE $S_1 \gets \tau(S_0)$ \COMMENT{Hold/defer: keep ambiguity as-is for this turn}
\STATE $S_2 \gets \delta(S_1, \lambda)$ \COMMENT{Dampen convergence}
\STATE $S_3 \gets \kappa(S_2, S_{\text{prev}})$ \COMMENT{Integrate with history (CPP)}
\STATE $S' \gets \pi(S_3, S_{\text{prev}})$ \COMMENT{Apply temporal persistence/decay}
\STATE $o \gets \Pi(S')$ \COMMENT{Non-destructive projection}

\RETURN $(S', o)$
\end{algorithmic}
\end{algorithm}

In this pipeline, $\tau$ is intentionally a no-op on weights: it marks
an explicit defer/hold decision instead of forcing an update.

\subsection{Computational Complexity}

\begin{itemize}
    \item \textbf{Conflict detection}: $O(|T| \times |\mathcal{M}|)$ where $|\mathcal{M}|$ 
          is the marker vocabulary size
    \item \textbf{Rule-based extraction}: $O(|T|)$ for segmentation
    \item \textbf{LLM-based extraction}: Dominated by LLM inference cost
    \item \textbf{State construction}: $O(|\mathcal{R}| \times d)$ for embedding
    \item \textbf{Merge deduplication}: $O(|\mathcal{R}|^2)$ for pairwise similarity
\end{itemize}

The paper reports asymptotic components only. It does not provide wall-clock, token,
memory, or end-to-end cost benchmarks.

\section{State Construction Audit}
\label{sec:experiments}

\subsection{Research Questions}

\begin{itemize}
    \item \textbf{RQ1}: Under the reported extraction procedures, do constructed states contain multiple positive-weight records?
    \item \textbf{RQ2}: Which reported categories yield multi-record states under rule-based vs. LLM-assisted extraction?
    \item \textbf{RQ3}: Can the rule-based detector be re-instantiated for the reported Japanese markers?
    \item \textbf{RQ4}: How consistent are results across different LLMs (ChatGPT, Gemini, Claude)?
\end{itemize}

\subsection{Test Set}

We constructed a mixed test set of 68 inputs spanning semantic ambiguity, explicit
conflict or ambivalence, and epistemic uncertainty. For rule-based extraction
(adversative, hedging), we include both English and Japanese sentences to illustrate
separate marker-set instantiation. For LLM-based extraction
(epistemic, lexical, structural), we use English only to isolate the contribution 
of our framework from the multilingual capabilities of the underlying LLMs.

\begin{table}[h]
\centering
\begin{tabular}{lcccc}
\toprule
\textbf{Category} & \textbf{Total} & \textbf{EN} & \textbf{JP} & \textbf{Method} \\
\midrule
Adversative & 20 & 10 & 10 & Rule-based \\
Hedging & 20 & 10 & 10 & Rule-based \\
Epistemic & 8 & 8 & -- & LLM-based \\
Lexical & 10 & 10 & -- & LLM-based \\
Structural & 10 & 10 & -- & LLM-based \\
\midrule
\textbf{Total} & \textbf{68} & \textbf{48} & \textbf{20} & \\
\bottomrule
\end{tabular}
\caption{Test set composition. Japanese sentences use the separately instantiated
rule-based detector described in Section~\ref{sec:framework}.}
\label{tab:testset}
\end{table}

\begin{figure}[ht]
\centering
\includegraphics[width=0.75\linewidth]{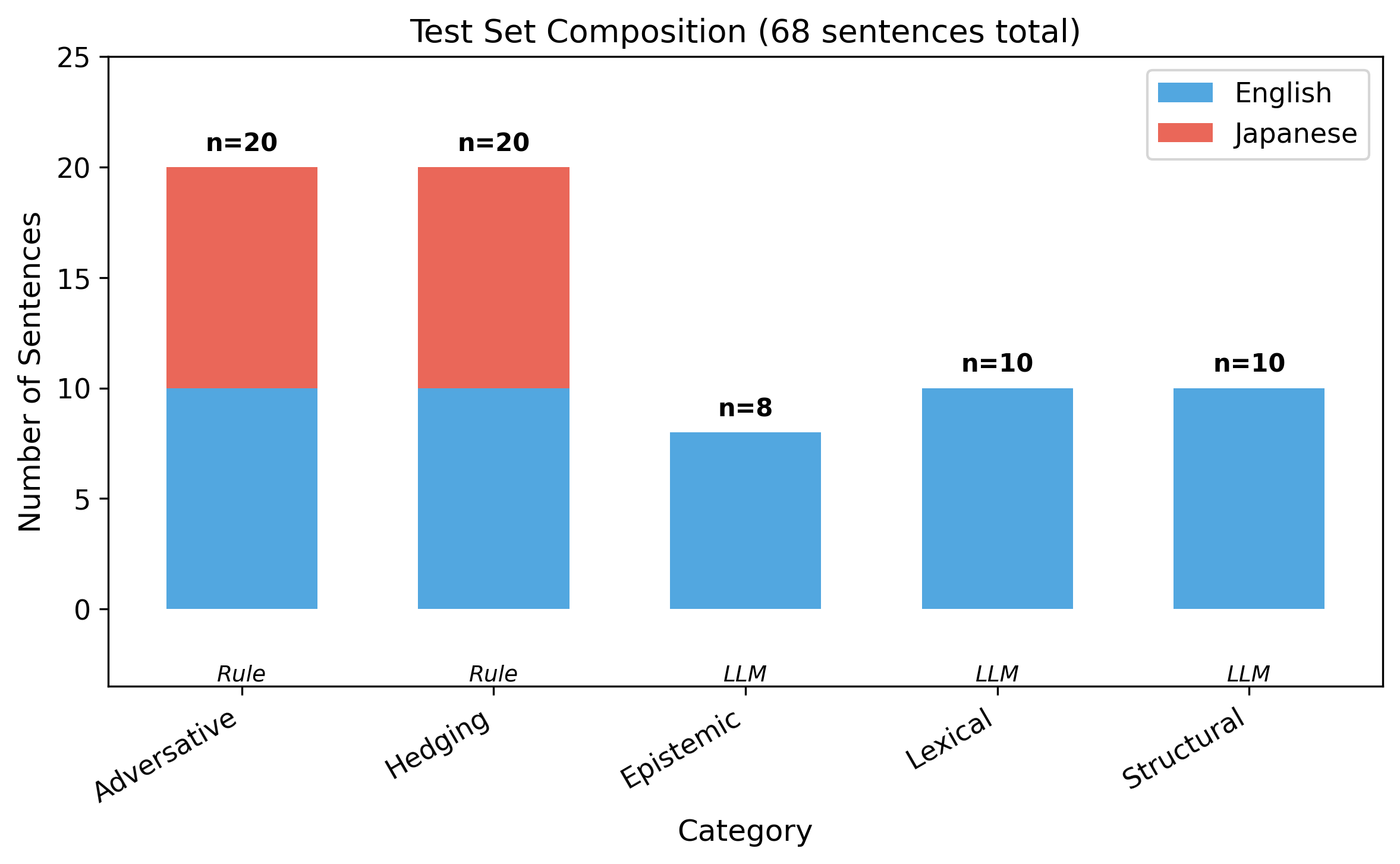}
\caption{Test set composition across five ambiguity categories (68 sentences total). 
Rule-based extraction targets adversative and hedging categories with both English 
and Japanese sentences. LLM-based extraction targets epistemic, lexical, and 
structural categories with English sentences only.}
\label{fig:testset}
\end{figure}

\textbf{Adversative} sentences contain explicit contradiction markers (``but'', ``however'', 
Japanese \textit{kedo}, \textit{demo}). Example: ``I want to quit my job, but I also don't want to quit.''

\textbf{Hedging} sentences contain uncertainty markers (``maybe'', ``perhaps'', 
Japanese \textit{kamoshirenai}). Example: ``Maybe I should apply for that position.''

\textbf{Epistemic} sentences contain belief markers (``I think'', ``I believe''). 
Example: ``I think I made the right choice.''

\textbf{Lexical ambiguity} sentences contain polysemous words. 
Example: ``I saw her duck.''

\textbf{Structural ambiguity} sentences have multiple parse trees. 
Example: ``I shot an elephant in my pajamas.''

\subsection{Metrics}

\begin{itemize}
    \item \textbf{State Size} $|S|$: Number of retained records in the state
    \item \textbf{State Entropy} $H(S)$: Shannon entropy of the weight distribution
    \item \textbf{Conflict Detection Rate}: Proportion of sentences where markers were detected
\end{itemize}

\noindent These are construction metrics over returned records. Record-weight entropy
measures the multiplicity and relative weights of those records, not their semantic
correctness, distinctness, or completeness. The $H=0$ condition denotes a single-record
state rather than an evaluated model baseline. Also,
$H_{\max} = \log_2 |S|$ depends on the number of positive-weight records;
$H_{\max} = 1.0$ bit applies only to binary ($|S|=2$) states, while states with
$|S| > 2$ can achieve $H > 1.0$ bits.

\subsection{Results: Rule-Based Extraction}

We first evaluate rule-based extraction alone. Table~\ref{tab:results-rule} shows
that the rules produce multi-record states for the reported adversative and hedging
patterns ($H \approx 1.0$), but do not produce multiple records for the epistemic,
lexical, and structural inputs ($H = 0$).

\begin{table}[h]
\centering
\begin{tabular}{lccc}
\toprule
\textbf{Category} & \textbf{$|S|$} & \textbf{$H(S)$} & \textbf{Method} \\
\midrule
Adversative & 2.10 & 1.037 & Rule-based \\
Hedging & 2.05 & 1.002 & Rule-based \\
Epistemic & 1.00 & 0.000 & Rule-based \\
Lexical & 1.00 & 0.000 & Rule-based \\
Structural & 1.00 & 0.000 & Rule-based \\
\bottomrule
\end{tabular}
\caption{Rule-based extraction results. Under the current rules, the epistemic,
lexical, and structural inputs require an auxiliary candidate source to produce
multi-record states.}
\label{tab:results-rule}
\end{table}

\noindent Note that Proposition~\ref{thm:noncollapse} requires $|\mathcal{R}(T)| > 1$
positive-weight retained records. For epistemic, lexical, and structural
categories, the current rule-based method returns one record
($|\mathcal{R}| = 1$), so the proposition's precondition is not met. The $H = 0$
results above are not counterexamples but rather motivate the hybrid $\phi$
that combines rule-based and LLM-based extraction.

\subsection{Results: LLM-Based Extraction}

To address the limitations of rule-based extraction, we audited archived
LLM-generated candidate sets from three models: ChatGPT, Gemini, and Claude.\footnote{Experiments used free-tier web interfaces in January 2026; exact model build IDs were not exposed.}
Each archived run proposed possible readings or epistemic alternatives for 28 English
sentences (8 epistemic + 10 lexical + 10 structural). The repository bundles
the prompts, saved outputs, and audit manifest used to reconstruct these
summaries from \texttt{prompts/}.

\begin{table}[h]
\centering
\begin{tabular}{lcccc}
\toprule
\textbf{Category} & \textbf{ChatGPT} & \textbf{Gemini} & \textbf{Claude} & \textbf{Mean} \\
\midrule
Epistemic $|S|$ & 3.00 & 3.00 & 3.75 & 3.25 \\
Epistemic $H(S)$ & 1.577 & 1.558 & 1.885 & 1.673 \\
\midrule
Lexical $|S|$ & 2.00 & 2.30 & 2.20 & 2.17 \\
Lexical $H(S)$ & 0.990 & 1.024 & 1.008 & 1.007 \\
\midrule
Structural $|S|$ & 2.20 & 2.10 & 2.10 & 2.13 \\
Structural $H(S)$ & 1.056 & 0.821 & 1.027 & 0.968 \\
\bottomrule
\end{tabular}
\caption{Archived LLM-assisted candidate summaries (English only) across three models.
All categories yield $H > 0$ in the audited outputs. The larger epistemic value reflects
more returned candidate records (e.g., assertion, hedge, or self-justification labels),
not adjudicated semantic coverage.}
\label{tab:results-llm}
\end{table}

\paragraph{Key Findings.}

Within the archived outputs, LLM-assisted extraction returns multi-record candidate
sets for all categories that the current rule-based methods do not cover:
\begin{itemize}
    \item \textbf{Epistemic}: Mean $|S| = 3.25$, $H = 1.673$ (the largest
          returned candidate sets)
    \item \textbf{Lexical}: Mean $|S| = 2.17$, $H = 1.007$ (multiple proposed readings)
    \item \textbf{Structural}: Mean $|S| = 2.13$, $H = 0.968$ (multiple proposed parses)
\end{itemize}

Within this archived set, Claude returned the most records on average
($|S| = 3.75$ for epistemic), while all three models yielded $H > 0$ for
100\% of test sentences.

\subsection{Combined Results}

Table~\ref{tab:results-combined} presents the combined results surface, combining rule-based
extraction (for adversative and hedging, including Japanese sentences) with
audited archived LLM extraction summaries (for epistemic, lexical, and structural,
English only).

\begin{table}[h]
\centering
\begin{tabular}{lcccc}
\toprule
\textbf{Category} & \textbf{N} & \textbf{$|S|$} & \textbf{$H(S)$} & \textbf{Method} \\
\midrule
Adversative & 20 & 2.10 & 1.037 & Rule (EN+JP) \\
Hedging & 20 & 2.05 & 1.002 & Rule (EN+JP) \\
Epistemic & 8 & 3.25 & 1.673 & LLM (EN) \\
Lexical & 10 & 2.17 & 1.007 & LLM (EN) \\
Structural & 10 & 2.13 & 0.968 & LLM (EN) \\
\midrule
\textbf{Overall} & \textbf{68} & \textbf{2.24} & \textbf{1.087} & Hybrid \\
\bottomrule
\end{tabular}
\caption{Combined results using the reported hybrid evidence surface. All categories
achieve $H > 0$, with overall mean entropy $H = 1.087$ bits. The overall row is
computed as the weighted mean across all 68 sentences, where LLM-based $H$
values represent the three-model average for each sentence in the archived set.}
\label{tab:results-combined}
\end{table}

Appendix~\ref{app:construction-audit-visuals} provides the archived-model and
combined construction-audit visual summaries. The tables above remain the primary
reader-visible evidence for the construction audit.

\subsection{Research Question Answers}

\textbf{(RQ1)} Under the reported extraction procedures, the constructed states contain
multiple positive-weight records, with mean record-weight entropy $H = 1.087$ bits
across the 68 inputs. This confirms that the reported procedures populate the typed
multi-record schema under the construction metrics defined above.

\textbf{(RQ2)} Rule-based extraction produces multi-record states for the reported
adversative and hedging patterns with $H \approx 1.0$. For the epistemic, lexical,
and structural categories in this paper's archived LLM-assisted set, the audited
outputs yield $H = 0.97$--$1.67$.

\textbf{(RQ3)} The rule-based conflict detector was separately instantiated for the
reported Japanese markers (\textit{kedo}, \textit{demo}, \textit{kamoshirenai}). This
illustrates re-instantiation only; equivalence and cross-lingual generalization were not tested.

\textbf{(RQ4)} In the archived three-model set, all three LLMs (ChatGPT, Gemini,
Claude) yield $H > 0$ for 100\% of English test sentences. Claude returns the
most records (mean $|S| = 3.75$ for epistemic), while Gemini is more
conservative ($|S| = 3.00$). Despite variation in $|S|$, all audited outputs
yield multi-record outputs on this dataset.

\subsection{Limitations}

The reported construction results have several limitations:

\begin{itemize}
    \item \textbf{Test set size}: 68 inputs across 5 categories provide a bounded
          construction audit; larger and independently annotated evaluation is needed.
    \item \textbf{LLM variability}: Different models return different numbers
          of records (Claude: 3.75, Gemini: 3.00 for epistemic), suggesting
          sensitivity to model choice.
    \item \textbf{Human evaluation}: We did not run a full human semantic-adequacy
          adjudication of the extracted records. The only additional check on
          the LLM-handled cases is a bounded 18-set current-API rerun: nine sets were
          consistent, eight were family-preserving reshapes, and one was divergent.
          This supports broad family-level stability but not human-labeled correctness.
    \item \textbf{Weight calibration}: Confidence weights from LLMs may not
          accurately reflect candidate plausibility.
    \item \textbf{Cross-lingual scope}: Japanese markers are separately instantiated
          only for rule-based extraction; equivalence and multilingual generalization
          are not evaluated.
\end{itemize}

\noindent\textbf{Note on LLM role:} LLMs serve here as archived sources of candidate enumeration; the theoretical claims of this paper concern the structure of the resulting states, not the capabilities of the underlying models. The $\phi$ mapping framework is model-agnostic and could be instantiated with another item source.
NRR is a derivation-and-evaluation framework, not a closed recipe; the same derivation lens supports alternative mapping/operator instantiations under fixed protocols with condition-bounded reporting.

\section{Operator Contract and Conformance Validation}
\label{sec:operator-validation}

Constructing a multi-record state is only the first step. Once records are
externalized, an update can still suppress weaker records and recreate premature
commitment at the state-maintenance layer. Phi therefore makes update behavior
inspectable through the record-weight entropy preservation criterion
\begin{equation}
H(\mathcal{O}(\mathcal{S})) \geq H(\mathcal{S}) - \epsilon,
\end{equation}
defined on admissible positive-total states. The criterion measures structural change
in the normalized record-weight distribution; it is not a semantic-quality metric or
a stand-alone record-identity/cardinality test.

The maintenance contract therefore exposes two distinct layers. Record-collection
behavior is read from each operator definition: one-to-one weight transformations keep
the input record collection, integration and persistence carry input records into a
union, and tagged-disjoint union has an explicit cardinality guarantee. The entropy
criterion separately detects concentration in normalized record weights. Neither layer
substitutes for the other: deleting the dominant record from weights
$[0.8,0.1,0.1]$ raises normalized entropy from approximately $0.922$ to $1.0$ bits, and
a weight permutation preserves entropy, even though record identity behavior differs.
The 2,740 measurements below audit the entropy layer; the stated record-retention
properties come from the declared operator definitions.

We evaluate the declared transition family on 580 constructed units: 180 single
states, 200 contradictory pairs, and 200 temporal pairs. Sweeps over uniform
subtraction, dampening, and calibration parameters yield 2,740 operator--state
measurements with $\epsilon=0.1$ bits. Table~\ref{tab:operator-main-summary}
summarizes the differential result; Appendix~\ref{app:operators} provides the operator
definitions, analytical properties, full methodology, figures, and extended table.
Every row is executed in the suite, but its inferential basis differs: dampening and
state calibration have analytic entropy results; deferred resolution is an identity
hold; integration and persistence carry records forward by definition while their
entropy rows remain tested conformance results; uniform subtraction is the designed
negative control.

\begin{table}[H]
\centering
\small
\begin{tabular}{llrr}
\toprule
\textbf{Transition/check} & \textbf{Setting} & \textbf{Measurements} & \textbf{Violations} \\
\midrule
Uniform subtraction ($\delta$ v1) & $b=0.05$ & 180 & 3 (1.7\%) \\
Uniform subtraction ($\delta$ v1) & $b=0.10$ & 180 & 11 (6.1\%) \\
Uniform subtraction ($\delta$ v1) & $b=0.20$ & 180 & 32 (17.8\%) \\
\midrule
Dampening ($\delta$ v2) & five $\lambda$ values & 900 & 0 (0\%) \\
State calibration ($\sigma_{\mathrm{state}}$) & four $b$ values & 720 & 0 (0\%) \\
Deferred resolution ($\tau$) & identity/hold & 180 & 0 (0\%) \\
CPP integration ($\kappa$) & contradictory pairs & 200 & 0 (0\%) \\
Persistence ($\pi$) & temporal pairs & 200 & 0 (0\%) \\
\bottomrule
\end{tabular}
\caption{Main operator-contract result across 2,740 measurements. Uniform
subtraction increasingly violates the $\epsilon$-preservation criterion as subtraction
grows; the tested declared transitions and checks show no violations under the reported
settings. Several zero rows are analytical or by construction, so this is executable
contract conformance with a discriminating negative control, not learned robustness.}
\label{tab:operator-main-summary}
\end{table}

\begin{figure}[H]
\centering
\includegraphics[width=0.85\textwidth]{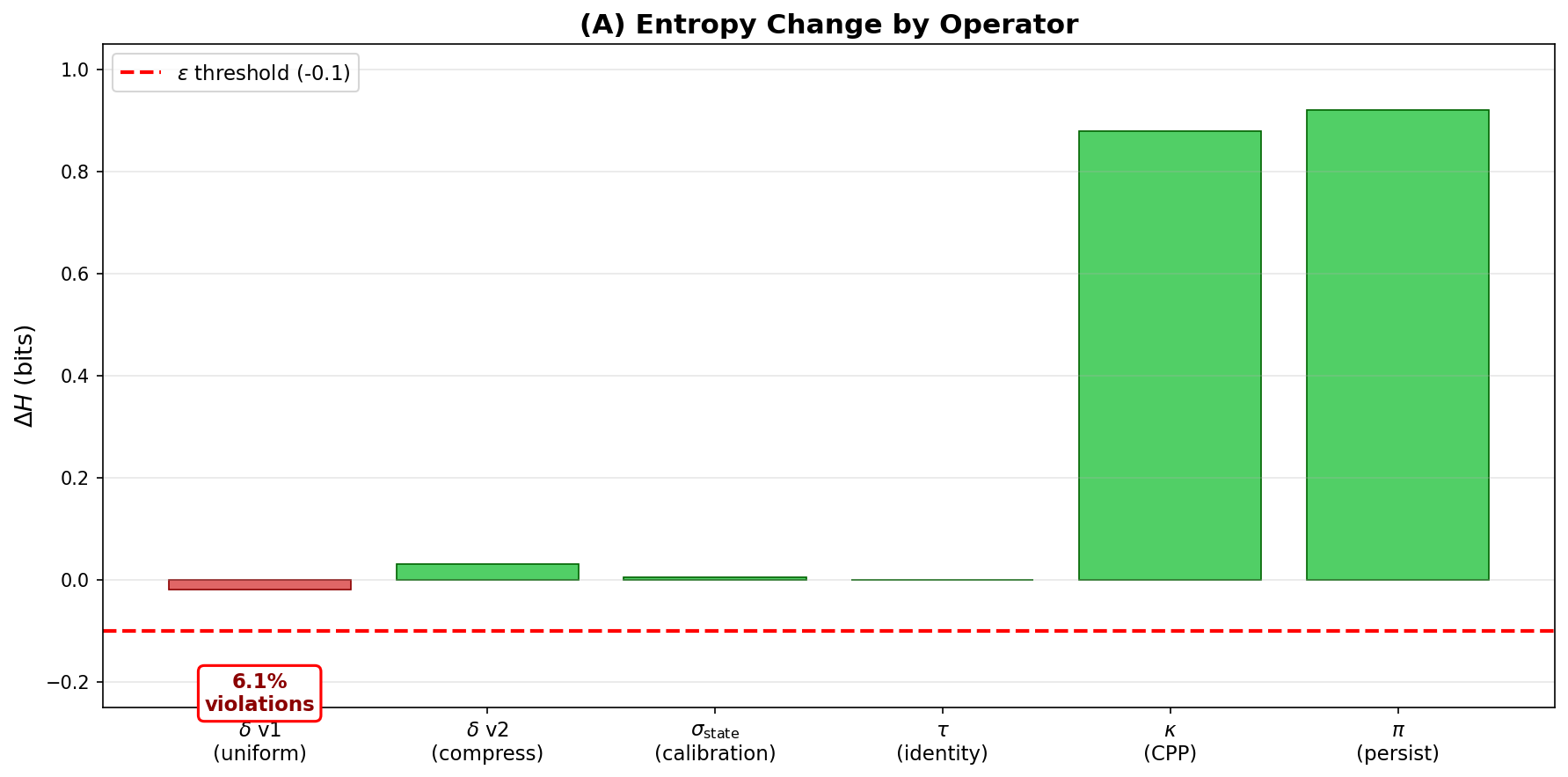}
\caption{Operator-conformance differential (representative $b = 0.10$ for uniform
subtraction). The red comparison crosses the $\Delta H=-\epsilon=-0.1$ threshold in
6.1\% of states at $b=0.10$ (1.7\% at $b=0.05$ and 17.8\% at $b=0.20$), whereas the
tested declared transitions and calibration/identity checks show 0\% violations.
Several zero bars are analytical or by construction; the affirmative result is an
executable update contract with a discriminating negative control, not learned
robustness or semantic-adequacy evidence.}
\label{fig:operators}
\end{figure}

The uniform-subtraction comparison provides the discriminating negative control. Its
violation rate rises monotonically with subtraction magnitude because a fixed decrement
disproportionately suppresses weaker records. By contrast, the tested declared family
shows 0\% violations in the reported settings. Several zero-violation rows are supported
analytically or by construction: state calibration preserves normalized entropy,
deferred resolution is an identity hold, dampening is entropy-nondecreasing on its
declared domain, and integration/persistence retain records by their definitions. The
result should therefore be read as executable contract conformance plus a negative
control, not as learned robustness.

This is Phi's direct link to premature commitment: after candidate records have been
externalized, the update contract states which records are carried forward, while the
entropy audit detects whether normalized weights are concentrated under the declared
criterion. The validation establishes entropy conformance for the tested operators and
states; the operator definitions separately provide their stated record-retention
properties. It does not establish that the records are semantically correct or that an
integrated system improves downstream decisions.

\section{Illustrative Application: Psychological Support}
\label{sec:application}

We briefly illustrate how $\phi$ applies to psychological support contexts, 
where ambivalence is common and premature resolution may be counterproductive.

\subsection{Motivation}

In support contexts, statements like ``I love them, but being with them hurts'' can
express ambivalence. An external multi-record state may help a system avoid selecting
one pole before its response policy has considered both; any user benefit is a downstream
hypothesis. Possible response failures include:
\begin{itemize}
    \item Diagnose the ambivalence as a ``problem''
    \item Offer resolution strategies (pros/cons lists, decision frameworks)
    \item Select one pole as primary
\end{itemize}

With $\phi$, the ambivalent statement maps to a state with two attitude records:
\begin{align*}
S = \{&(\text{``I love them''}, \text{pre-adv}, 0.96, \{\text{conflict}: \text{True}\}), \\
      &(\text{``being with them hurts''}, \text{post-adv}, 0.96, \{\text{conflict}: \text{True}\})\}
\end{align*}

Both records are preserved with equal weight. A downstream response policy could use
them to acknowledge coexistence rather than forcing resolution; that response effect is
not evaluated here.

\subsection{Limitations}

This application is \textbf{illustrative, not validated}. Whether NRR-based systems 
produce therapeutically beneficial outcomes requires:
\begin{itemize}
    \item Clinical trials with appropriate ethical oversight
    \item Comparison with established therapeutic approaches
    \item Long-term outcome measurement
\end{itemize}

NRR is a computational architecture, not a therapeutic intervention.

\section{Discussion}
\label{sec:discussion}

\subsection{Implications for LLM Architecture}

Natural-language inputs can remain underspecified under the context available at a
given turn. When more than one reading or state item remains live, selecting one for
output can discard alternatives that later evidence may support. The role of $\phi$ is
to externalize the extracted items into a typed state and thereby separate retained
state from output selection.

The present work establishes both the typed representational option and its maintenance
criterion at the system level. The construction audit shows that the reported procedures
populate explicit states, while the operator audit shows that update-rule choice is
detectably consequential: the declared family conforms to the record-weight invariant
in the tested settings, whereas uniform subtraction increasingly suppresses weaker
records. An integrated system could use this external state across output boundaries
and revise it as evidence arrives while still emitting a singular response when
required. End-to-end commitment behavior remains a separate empirical question.

\subsection{Distinction from Prompt-Based Enumeration}

A natural question arises: if LLMs can enumerate multiple interpretations when 
prompted, why is $\phi$ necessary? Note that prompting an LLM to enumerate 
interpretations produces a textual list but does not by itself provide a typed state
contract, persistence rule, or operator interface across turns. The present paper does
not infer from that list what the base model simultaneously represents internally.
Instead, $\phi$ externalizes the extracted items into an explicit state $S$
that a system can preserve and transform through declared NRR operators. The distinction
is therefore between \textit{enumerating} interpretations and \textit{constructing an
operator-addressable retained state}. Local judgment can still occur inside the system;
the point is that retention itself need not be implemented as repeated full branchwise
comparative evaluation during evidence accumulation.

\subsection{Limitations}

\paragraph{Marker Coverage.}
The conflict marker taxonomy is not exhaustive. Languages beyond English and 
Japanese require additional marker sets.

\paragraph{LLM Dependence.}
LLM-based extraction depends on external model availability and introduces 
non-determinism. Rule-based extraction is reproducible but limited in coverage.

\paragraph{Weight Calibration.}
Current weight assignment ($w = \text{conf} \times (1 + \beta)$) is heuristic. 
Learned weight calibration could improve entropy properties.

\paragraph{Evaluation Scope.}
The 68-sentence test set covers common ambiguity patterns but may not represent 
all real-world inputs.

\subsection{Future Work}

\begin{itemize}
    \item \textbf{Native Non-Collapsing Inference}: Integrate $\phi$ into LLM 
          architectures rather than as a preprocessing step
    \item \textbf{Learned Conflict Detection}: Train classifiers for marker 
          detection rather than pattern matching
    \item \textbf{Cross-Lingual Extension}: Develop marker taxonomies for 
          additional languages
    \item \textbf{Human Evaluation}: Assess perceived quality of non-collapsing 
          responses in user studies
\end{itemize}

\paragraph{Reproducibility note.}
We release deterministic code, fixed settings, and tracked artifacts for the
rule-based mapping and operator-validation reruns. For the main-text LLM table,
the package releases archived prompts, saved outputs, a sentence-level manifest,
and a local audit script that reconstruct the reported summaries rather than
replaying live model calls. The full environment specification and artifact map
are provided in \texttt{nrr-phi/reproducibility.md}.

\subsection{Broader Impact and Risk Mitigation}

The illustrative psychological-support framing in Section~\ref{sec:application}
raises asymmetric-risk concerns: a system that preserves ambiguity can still be
misread as therapeutic guidance, can surface multiple harmful interpretations
without adequate guardrails, or can encourage over-reliance in emotionally
sensitive settings. The intended contribution of Phi is therefore architectural
and diagnostic, not clinical. Risk mitigation for this line includes keeping the
application framing explicitly non-therapeutic, avoiding claims of treatment
benefit, pairing any deployment-facing use with human oversight and domain
policies, and treating downstream evaluation in sensitive settings as a separate
ethics-reviewed task rather than as a consequence of the present results.

\section{Conclusion}
\label{sec:conclusion}

This paper presented a typed external text-to-state interface and a declared
state-maintenance contract for Non-Resolution Reasoning. The mapping transforms
extracted items into operator-addressable $(v,c,w,m)$ records; the update layer makes
two properties inspectable rather than implicit: which records a declared operator
carries forward and how their normalized weights change.

The main validation applies the contract to 580 constructed states/pairs and 2,740
operator--state measurements. The tested non-violating transitions and checks show 0\%
record-weight entropy violations in the reported settings, while uniform subtraction
violates the criterion at 1.7\%, 6.1\%, and 17.8\% as subtraction increases. This
differential result supplies both an executable conformance test and a negative control
for entropy-concentrating suppression at the retained-state layer.

The 68-input construction audit plays a supporting role: the reported rule-based and
archived LLM-assisted procedures populate the typed schema with multiple positive-weight
records, with mean record-weight entropy $H=1.087$ bits. That structural result does not
adjudicate semantic correctness, and maintenance-layer conformance does not by itself
establish improved downstream decisions. Those boundaries separate the present
engineering contract from future end-to-end evaluation without weakening what the
contract already provides.

Phi turns retained-state maintenance into a typed, executable, and falsifiable
interface for inspecting record carry-forward and normalized-weight concentration
before a declared output policy commits.


\appendix

\section{State Construction Audit Visuals}
\label{app:construction-audit-visuals}

These supporting figures visualize the archived construction audit summarized by
the main-text tables. They do not rank providers or adjudicate returned meanings.

\begin{figure}[ht]
\centering
\includegraphics[width=0.85\linewidth]{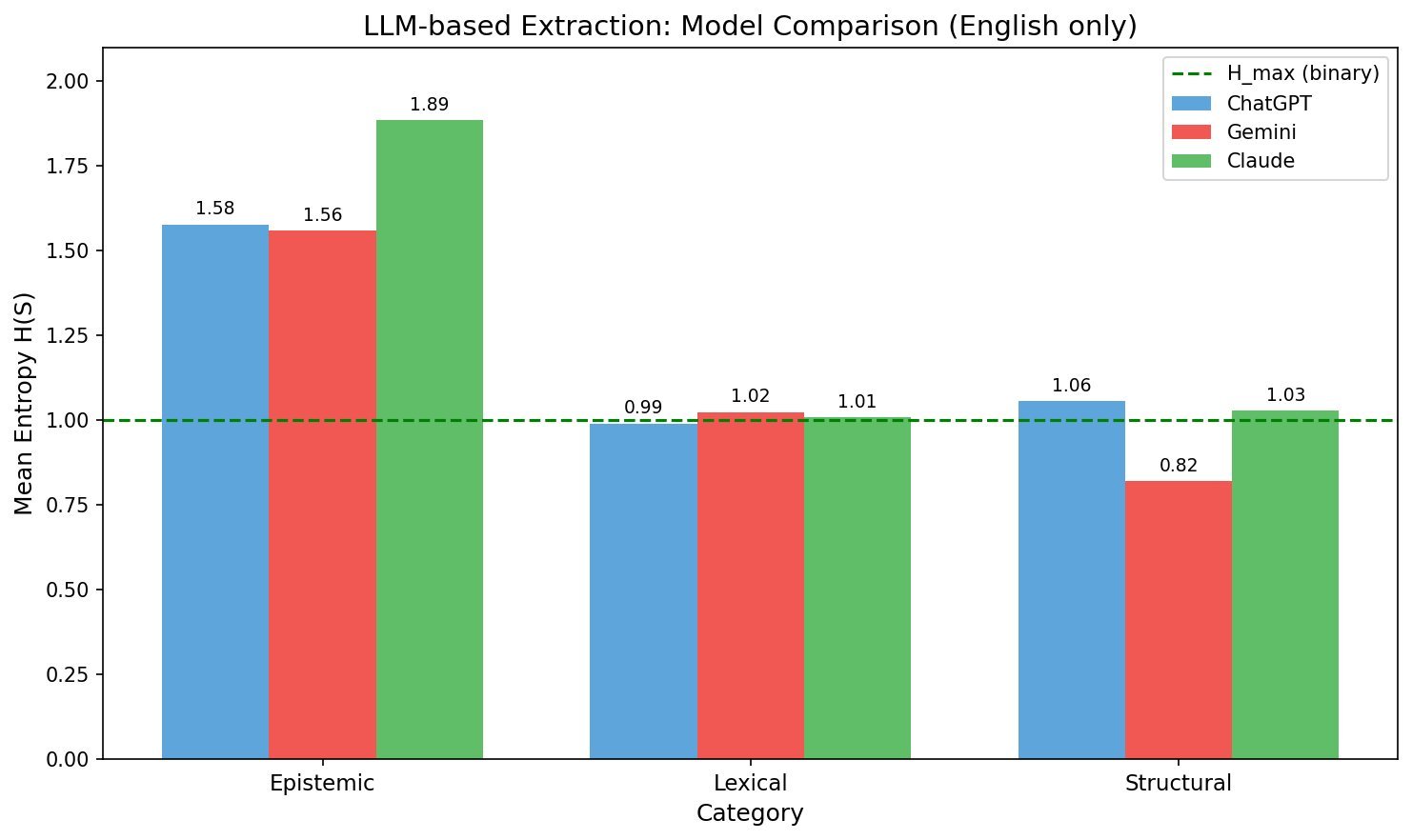}
\caption{Archived LLM-assisted construction summaries across three models
(ChatGPT, Gemini, Claude) for English sentences. The dashed line indicates
$H_{\max}=1.0$ for two-record states. All audited outputs yield $H>0$ across the
reported categories; the visual reports returned record-weight distributions, not
provider quality or adjudicated semantic coverage.}
\label{fig:llm-comparison}
\end{figure}

\begin{figure}[ht]
\centering
\includegraphics[width=\linewidth]{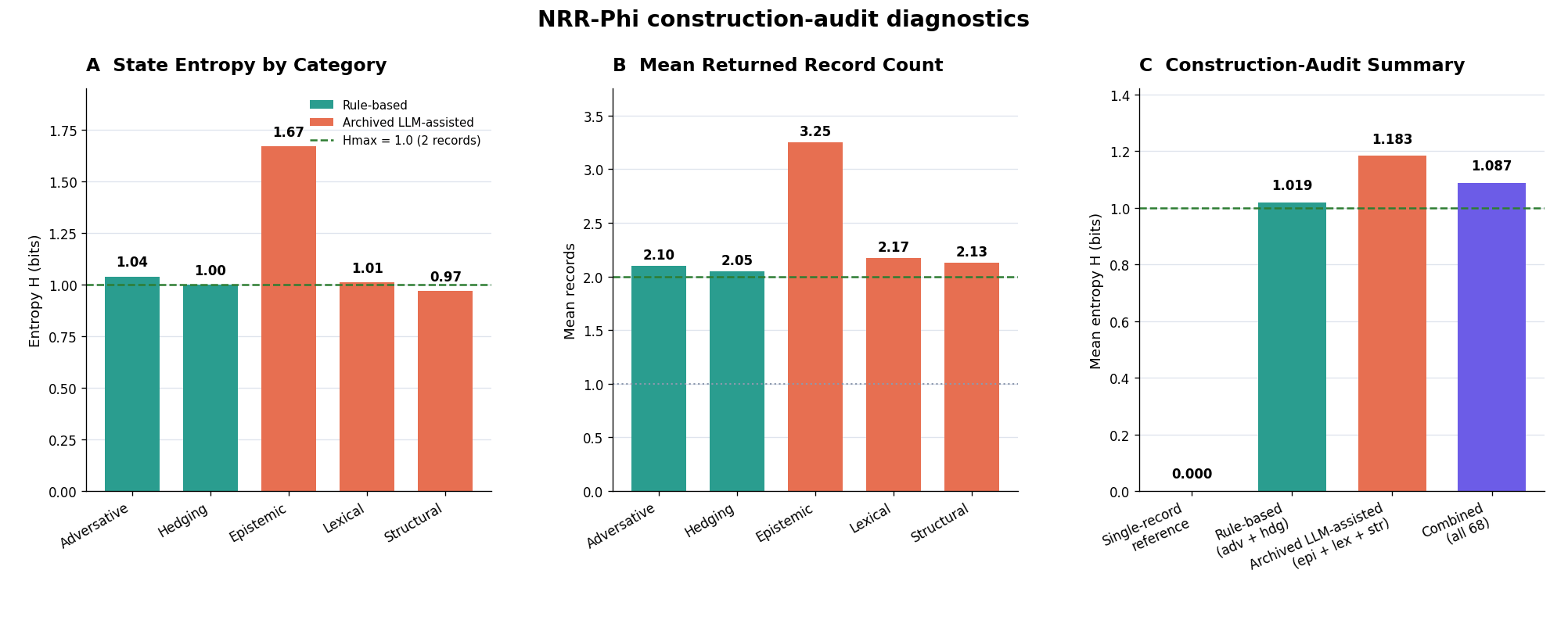}
\caption{Construction-audit summary. \textbf{(A)} State entropy by category and
extraction method. Rule-based extraction achieves $H\approx1.0$ for adversative and
hedging; archived LLM-assisted extraction achieves $H>0$ for all reported categories.
\textbf{(B)} Mean returned record count by category. \textbf{(C)} Stipulated
single-record reference ($H=0$), rule-based ($H=1.019$), archived LLM-assisted
($H=1.183$), and combined ($H=1.087$). These are schema-population diagnostics, not
semantic-adequacy or end-to-end performance results.}
\label{fig:results-summary}
\end{figure}

\section{Conflict Marker Taxonomy}
\label{app:markers}

\begin{table}[h]
\centering
\small
\setlength{\tabcolsep}{4pt}
\begin{tabular}{llll}
\toprule
\textbf{Category} & \textbf{English} & \textbf{Japanese (romanized)} & \textbf{Context Label} \\
\midrule
\multicolumn{4}{l}{\textit{Explicit Contradiction}} \\
Adversative & but, however, yet, although & \textit{kedo, demo, shikashi, daga} & \texttt{pre-adv}, \texttt{post-adv} \\
Contrastive & on the other hand, whereas & \textit{ippou de, hanmen} & \texttt{contrast-A}, \texttt{contrast-B} \\
Concessive & even though, despite & \textit{nimo kakawarazu} & \texttt{concede}, \texttt{main} \\
\midrule
\multicolumn{4}{l}{\textit{Implicit Uncertainty}} \\
Hedging & maybe, perhaps, might & \textit{kamoshirenai, tabun} & \texttt{hedge-scope} \\
Epistemic & I think, I believe, it seems & \textit{to omou, ki ga suru} & \texttt{epistemic-stance} \\
Modal & could, would, should & \textit{beki, hazu, darou} & \texttt{modal-world} \\
\midrule
\multicolumn{4}{l}{\textit{Structural Ambiguity}} \\
Coordination & both X and Y, either X or Y & \textit{mo...mo, ka...ka} & \texttt{coord-A}, \texttt{coord-B} \\
Scope & all...not, every...some & -- & \texttt{wide-scope}, \texttt{narrow-scope} \\
\bottomrule
\end{tabular}
\caption{Conflict marker taxonomy (English and Japanese)}
\label{tab:markers-full}
\end{table}

\section{LLM Prompt Template}
\label{app:prompt}

The following prompt template is used for LLM-based interpretation extraction:

\begin{verbatim}
Given the text: "[T]"

[If conflict markers detected: 
"Note: This text contains potential ambiguity markers."]

List ALL possible interpretations as distinct meanings.
For each interpretation, provide:
1. The interpretation (a clear restatement of one possible meaning)
2. The context/condition under which this interpretation holds
3. Confidence weight from 0.0 to 1.0

Format each as:
INTERP: [interpretation]
CONTEXT: [context]
CONFIDENCE: [0.0-1.0]
---
\end{verbatim}

\section{Sample Outputs}
\label{app:samples}

This appendix provides representative examples of how $\phi$ processes different 
ambiguity types. We show one example each from lexical ambiguity (LLM-based), 
structural ambiguity (LLM-based), and adversative (rule-based with Japanese).

\subsection*{Example 1: Lexical Ambiguity (LLM-based)}

\textbf{Input}: ``I saw her duck.''

\textbf{ChatGPT Output}:
\begin{verbatim}
{
  "id": "lex_en_01",
  "interpretations": [
    {
      "meaning": "I observed her pet duck (the bird).",
      "context": "noun-reading",
      "confidence": 0.5
    },
    {
      "meaning": "I saw her lower her head quickly.",
      "context": "verb-reading",
      "confidence": 0.5
    }
  ]
}
\end{verbatim}

\textbf{State constructed by $\phi$}: $|S| = 2$, $H(S) = 1.00$ bits

\subsection*{Example 2: Structural Ambiguity (LLM-based)}

\textbf{Input}: ``I shot an elephant in my pajamas.''

\textbf{ChatGPT Output}:
\begin{verbatim}
{
  "id": "str_en_01",
  "interpretations": [
    {
      "meaning": "The speaker was wearing pajamas 
                  when they shot the elephant.",
      "context": "PP attaches to subject",
      "confidence": 0.7
    },
    {
      "meaning": "The elephant was inside the speaker's 
                  pajamas when it was shot.",
      "context": "PP attaches to object",
      "confidence": 0.3
    }
  ]
}
\end{verbatim}

\textbf{State constructed by $\phi$}: $|S| = 2$, $H(S) = 0.88$ bits

\subsection*{Example 3: Japanese Adversative (Rule-based)}

\textbf{Input}: ``\textit{Yametai kedo yametakunai}'' 
(``I want to quit, but I also don't want to quit.'')

\textbf{Rule-based segmentation}: Marker ``\textit{kedo}'' detected.

\textbf{Segments extracted}:
\begin{itemize}
    \item Segment 1: ``\textit{yametai}'' (``I want to quit'') --- context: \texttt{pre-adv}
    \item Segment 2: ``\textit{yametakunai}'' (``I don't want to quit'') --- context: \texttt{post-adv}
\end{itemize}

\textbf{Weights}: Both segments receive $w = 0.5$ (equal weight for adversative).

\textbf{State constructed by $\phi$}: $|S| = 2$, $H(S) = 1.00$ bits

\vspace{1em}
\noindent These examples illustrate the difference between prompted enumeration
(which proposes semantic readings) and rule-based extraction (which segments at
markers). Both produce multi-record states in these examples; this structural property
does not establish that the records are semantically correct or exhaustive.

\section{NRR Operator Design Principles and Definitions}
\label{app:operators}

This appendix details design principles and formal definitions for NRR's
state-transition machinery. We distinguish (i) core transition operators
used for state evolution and (ii) metadata functions used for representation
management. While the $\phi$ mapping (main text) handles text-to-state
transformation, these components govern state-to-state processing.

\subsection{Design Principles for Non-Collapsing Operators}
\label{app:principles}

Transforming NRR states risks destroying the record multiplicity that NRR preserves.
We identify four operator-design requirements. Each concrete operator must still be
proved or tested against the quantitative record-weight entropy criterion; listing a
requirement does not by itself establish preservation.

\subsubsection{Principle 1: Relative Structure Preservation}

\begin{principle}[Relative Structure Preservation]
\label{prin:relative}
An operator should preserve the proportional relationships between
record weights when its purpose is not redistribution. If $w_i / w_j = r$ before the operation,
then $w'_i / w'_j \approx r$ after.
\end{principle}

\textbf{Rationale}: Entropy depends on relative, not absolute, weight 
values. Transformations that preserve weight \textit{ratios} maintain 
the entropy structure. Note that ratio preservation is a \textit{sufficient} 
condition for non-collapse, but not strictly \textit{necessary}---operators 
such as dampening ($\delta$) preserve non-collapse by being entropy-nondecreasing, 
even when ratios are not strictly preserved.

\textbf{Implementation}: Use multiplicative transformations 
(e.g., $w'_i = \lambda w_i$) or proportional adjustments 
(e.g., $w'_i = w_i - b \cdot w_i / \max w$) rather than uniform 
additive changes.

\subsubsection{Principle 2: Scale Equivariance}

\begin{principle}[Scale Equivariance]
\label{prin:scale}
An operator's effect should be independent of the absolute scale 
of weights. Scaling all weights by a constant should yield the 
same entropy change.
\end{principle}

\textbf{Rationale}: Since weights are non-normalized, absolute 
magnitudes are arbitrary. Operations should depend only on relative 
structure.

\textbf{Implementation}: Design operators such that 
$\mathcal{O}(\lambda \mathcal{S}) \sim \lambda \mathcal{O}(\mathcal{S})$ 
for scalar $\lambda > 0$. This property is verifiable analytically 
from the operator definition.

\subsubsection{Principle 3: Contradiction Non-Destruction}

\begin{principle}[Contradiction Non-Destruction]
\label{prin:contradiction}
When merging states with conflicting records, preserve
both rather than eliminating one by default. Their conflict relation is retained
structure, not by itself an operator error.
\end{principle}

\textbf{Rationale}: This implements the Contradiction-Preservation
Principle from NRR foundations. Deleting one conflicting record removes a retained
candidate; whether either record is semantically correct requires separate evaluation.

\textbf{Implementation}: Union with conflict tagging rather than 
replacement or selection.

\subsubsection{Principle 4: Temporal Persistence}

\begin{principle}[Temporal Persistence]
\label{prin:temporal}
Records from previous states should persist (with decay)
rather than being overwritten, enabling reactivation when context 
shifts.
\end{principle}

\textbf{Rationale}: Context may shift to favor previously dormant
records. Complete overwriting destroys a candidate that may become relevant.

\textbf{Implementation}: Decayed union of current and previous states.

\subsubsection{The Record-Weight Entropy Preservation Criterion}

The principles converge on a single quantitative criterion.

\begin{definition}[Record-Weight Entropy Loss]
The record-weight entropy loss induced by operator $\mathcal{O}$ is:
\begin{equation}
L(\mathcal{O}, \mathcal{S}) = H(\mathcal{S}) - H(\mathcal{O}(\mathcal{S}))
\end{equation}
Positive $L$ indicates a decrease in record-weight entropy; negative $L$ indicates
an increase. This metric does not measure semantic information gain or loss.
\end{definition}

\begin{definition}[Record-Weight Entropy Preservation Criterion]
An operator $\mathcal{O}$ satisfies the $\epsilon$-preservation law if:
\begin{equation}
H(\mathcal{O}(\mathcal{S})) \geq H(\mathcal{S}) - \epsilon
\end{equation}
for all admissible positive-total states $\mathcal{S}$ whose transformed state
also has positive total activation, and small threshold $\epsilon$.
Equivalently, $L(\mathcal{O}, \mathcal{S}) \leq \epsilon$.
\end{definition}

The four principles are design constraints, not a general theorem that implies the
criterion for every operator class. Satisfaction of the criterion must be established
analytically or empirically for each declared operator and admissible domain. Passing
the criterion does not replace a separate inspection of record identity or cardinality;
entropy can increase after record deletion and is invariant to weight permutation.

\subsection{Operator Definitions}
\label{app:operator-defs}

We define four core transition operators included in the executable conformance suite
(\S\ref{app:validation}), plus one state-level calibration check and three metadata
functions. Their record carry-forward properties come from the definitions; the suite
separately checks the record-weight entropy criterion.

\subsubsection{Core Transition Operators (Conformance Checked)}

The following core transition operators are exercised on 180 single states,
200 contradictory pairs, and 200 temporal pairs constructed from three archived LLM
surfaces.

\paragraph{Dampening ($\delta$).}
Compresses weights toward the mean, reducing dominance of any single record:

\begin{equation}
\delta(\mathcal{S}) = \{(v_i, c_i, w_i(1-\lambda) + \bar{w}\lambda, m_i)\}
\end{equation}

where $\bar{w} = \frac{1}{n}\sum_j w_j$ and $\lambda\in[0,1]$.
On this convex-mixture domain, nonnegative weights remain nonnegative and a
positive-total state remains in the entropy-evaluation domain. Although it does not
preserve weight ratios, dampening is entropy-nondecreasing:
$H(\delta_{\lambda}(\mathcal{S})) \geq H(\mathcal{S})$. Strict increase occurs
exactly when $\lambda>0$ and the normalized input weight vector is nonuniform;
$\lambda=0$ and uniform inputs give equality. Thus it satisfies the
$\epsilon$-preservation law. The executable suite reports 0\% criterion violations across all tested settings.

\paragraph{State-level Calibration ($\sigma_{\mathrm{state}}$).}
Applies a uniform multiplicative scaling to all weights (state-level calibration check):

\begin{equation}
\sigma_{\mathrm{state}}(\mathcal{S}) = \left\{
  \left(v_i, c_i, w_i - b \cdot \frac{w_i}{\max_j w_j}, m_i\right)
\right\}
\end{equation}

For $0 \leq b < \max_j w_j$, this operation multiplies all weights within a state by
the same positive factor $\left(1 - b/\max_j w_j\right)$.
Therefore, after normalization for entropy evaluation, the induced probability
distribution is unchanged and $\Delta H = 0$. We retain this as a
state-level entropy calibration check, not as an entropy-changing update rule.
For fixed absolute $b$, it is not operator-scale-equivariant under rescaling of the
input state; only its within-state normalized entropy outcome is invariant on the
admissible domain.
This $\sigma_{\mathrm{state}}$ is separate from item-level strengthen operators
used in later implementation papers.

\paragraph{Deferred Resolution ($\tau$).}
Identity during processing; resolution only at output boundary:

\begin{equation}
\tau(\mathcal{S}) = \mathcal{S}
\end{equation}

Satisfies the record-weight entropy criterion by doing nothing until output is required.
Serves as the identity check in the executable conformance suite.

\paragraph{CPP Integration ($\kappa$).}
Implements Principle 3 (Contradiction Non-Destruction):

\begin{equation}
\kappa(\mathcal{S}, \mathcal{S}') = \mathcal{S} \cup \mathcal{S}' 
\text{ with conflict tags}
\end{equation}

Contradictory records are preserved by definition, not eliminated. The entropy check is
executed on 200 contradictory pairs and reports 0\% criterion violations.

\paragraph{Persistence ($\pi$).}
Implements Principle 4 (Temporal Persistence):

\begin{equation}
\pi(\mathcal{S}_t, \mathcal{S}_{t-1}) = \mathcal{S}_t \cup 
\{(v_i, c_i, \gamma w_i, m_i) : (v_i, c_i, w_i, m_i) \in \mathcal{S}_{t-1}\}
\end{equation}

Previous records persist by definition with decayed activation ($\gamma \in (0,1)$
is the decay coefficient). The entropy check is executed on 200 temporal pairs and
reports 0\% criterion violations.

\subsubsection{Metadata Functions (Representation Utilities)}

The following three functions preserve weights by construction and primarily
augment representation metadata rather than perform transition dynamics.
They are included as extensible utilities over retained records.

\paragraph{Positioning ($\rho$).}
Assigns temporal coordinates to distinguish identical tokens across 
contexts:

\begin{equation}
\rho(\mathcal{S}) = \{(v_i, (c_i, t), w_i, m_i)\}
\end{equation}

Implements contextual identity tracking without affecting weights. 
Preserves entropy by construction since no weights are modified.

\paragraph{Abstraction ($\alpha$).}
Augments state with relational structure:

\begin{equation}
\alpha(\mathcal{S}) = \{(v_i, c_i, w_i, m_i \cup \{\text{relations}:R_i\})\}
\end{equation}

where $R_i = \{d(v_i, v_j) : j \neq i\}$ captures geometric relations. 
Preserves entropy by construction since no weights are modified.

\paragraph{Invariance ($\iota$).}
Marks records originating from the same symbol across multiple
contexts:

\begin{equation}
\iota(\mathcal{S}_1, \ldots, \mathcal{S}_n)
= \biguplus_{j=1}^{n}\bigl(\{j\}\times\mathcal{S}_j\bigr)
\text{ with cross-context ID}
\end{equation}

where $\biguplus$ is a tagged disjoint (equivalently, record-preserving multiset)
union: the source-state tag $j$ prevents coincident records from being identified.
Each record $s_i \in \mathcal{S}_j$ also receives metadata marking it as derived
from the same underlying symbol (e.g., the word ``bank'' appearing in financial
vs.\ riverbank contexts). By construction, $\iota$ retains all records from all
input contexts and
$|\iota(\mathcal{S}_1, \ldots, \mathcal{S}_n)| = \sum_{j}|\mathcal{S}_j|$.
This cardinality guarantee does not imply that the normalized entropy of the joined
state is at least that of every input state; no unconditional entropy-ordering claim
is made for $\iota$.

\subsubsection{Non-Destructive Projection}

For output generation:

\begin{equation}
\Pi(\mathcal{S}) = \arg\max_i w_i, \quad \mathcal{S}' = \mathcal{S}
\end{equation}

The projection selects a singular output while leaving the specified external state
unchanged. This definition makes no claim about a base model's hidden state.
Note that $\Pi$ is an \textit{output} mechanism, not a processing 
operator; it is not subject to the design principles, since its purpose 
is precisely to select a single record for external communication.

\subsection{Theoretical Properties}
\label{app:properties}

\subsubsection{Operator--Data Structure Alignment}

Different components require different data structures for validation. 
Weight-transforming transitions ($\delta$, $\sigma_{\mathrm{state}}$, $\tau$) operate on 
individual states and are naturally validated on single states. In contrast, 
$\kappa$ (contradiction integration) and $\pi$ (temporal persistence) are 
\textit{binary} operators that take two states as input---validating 
them on single states would be meaningless, as it would fail to 
exercise their core function. Metadata functions $\alpha$ and $\rho$ neither
modify weights nor change the record collection and therefore preserve entropy
by construction. The tagged-union utility $\iota$ preserves every input record
and its weight, but changes the normalized mixture; record retention alone does
not imply an entropy ordering against each input.

\subsubsection{Composition Behavior}

If every member of a $k$-operator chain has independently established
$H(\mathcal{O}(\mathcal{S})) \geq H(\mathcal{S}) - \epsilon$, the chain has the
conservative bound $\Delta H \geq -k\epsilon$. For the specific chain
$\delta \circ \tau \circ \rho$ under the definitions here and
$\lambda\in[0,1]$, $\delta$ moves weights toward the mean and therefore has
$\Delta H \geq 0$, while $\tau$ and $\rho$ leave
weights unchanged. Consequently this named chain has $\Delta H \geq 0$; the earlier
generic $-k\epsilon$ bound is not needed to establish its behavior.

\subsubsection{Scale and Entropy Behavior}

Principle 2 (Scale Equivariance) requires that scaling all weights by
a constant $c > 0$ commute with the operator up to the same factor. Since
$H(\mathcal{S})$ depends only on normalized probabilities 
$p_i = w_i / \sum_j w_j$, and scaling all $w_i$ by $c$ leaves 
$p_i$ invariant, entropy invariance alone is weaker than operator equivariance.
The operators have the following properties:

\begin{itemize}
  \item \textbf{$\sigma_{\mathrm{state}}$ (State-level calibration check)}:
    for fixed absolute $b$,
    \begin{equation*}
    \begin{aligned}
    \sigma_{\mathrm{state}}(c\mathcal{S})_i
    &= c w_i - b\,w_i/\max w, \\
    c\,\sigma_{\mathrm{state}}(\mathcal{S})_i
    &= c w_i - cb\,w_i/\max w.
    \end{aligned}
    \end{equation*}
    These are unequal unless $c=1$ (or $b=0$), so the operator is not
    scale-equivariant. On its admissible domain it nevertheless rescales all records
    within a state uniformly and therefore has $\Delta H=0$.

  \item \textbf{$\delta$ (Dampening)}:
    for $\lambda\in[0,1]$,
    $\delta(c\mathcal{S})_i = c w_i(1-\lambda) + c\bar{w}\lambda
    = c[w_i(1-\lambda) + \bar{w}\lambda] = c \cdot \delta(\mathcal{S})_i$.
    Exact scale equivariance on the admissible domain.

  \item \textbf{$\tau$ (Identity)}: Trivially equivariant.

  \item \textbf{$\kappa$ (CPP)}: Union of sets; scaling both input states 
    by $c$ scales all weights in the output by $c$. Equivariant.

  \item \textbf{$\pi$ (Persistence)}: 
    $\pi(c\mathcal{S}_t, c\mathcal{S}_{t-1})$ scales all 
    weights (including decayed ones) by $c$. Equivariant.

  \item \textbf{$\rho$, $\alpha$, $\iota$}: These operators do not modify 
    weights, so scale equivariance is trivially satisfied.
\end{itemize}

With the exception of the fixed-$b$ state-calibration check, all listed transitions
and metadata functions are scale-equivariant as defined above.
$\sigma_{\mathrm{state}}$ is retained only as an entropy-invariance calibration check.

\subsection{Operator--Principle Correspondence}
\label{app:correspondence}

Table~\ref{tab:correspondence} summarizes which principles each 
operator satisfies. P1 = Relative Structure Preservation, 
P2 = Scale Equivariance, P3 = Contradiction Non-Destruction, 
P4 = Temporal Persistence. Operators marked with \textsuperscript{\ddag} are included
in the executable conformance suite (\S\ref{app:validation}).

\begin{table}[h]
\centering
\begin{tabular}{lcccc}
\toprule
\textbf{Operator} & \textbf{P1} & \textbf{P2} & \textbf{P3} & \textbf{P4} \\
\midrule
\multicolumn{5}{l}{\textit{Conformance-checked core transitions:}} \\
$\delta$ (Dampening)\textsuperscript{\ddag}     & $\approx$\textsuperscript{*} & \checkmark & -- & -- \\
$\sigma_{\mathrm{state}}$ (Calibration)\textsuperscript{\ddag} & \checkmark & --\textsuperscript{**} & -- & -- \\
$\tau$ (Hold/Defer)\textsuperscript{\ddag}      & \checkmark & \checkmark & -- & -- \\
$\kappa$ (CPP)\textsuperscript{\ddag}           & \checkmark & \checkmark & \checkmark & -- \\
$\pi$ (Persistence)\textsuperscript{\ddag}      & \checkmark & \checkmark & -- & \checkmark \\
\midrule
\multicolumn{5}{l}{\textit{Metadata functions (representation utilities):}} \\
$\rho$ (Positioning)\textsuperscript{\S}        & \checkmark & \checkmark & -- & -- \\
$\alpha$ (Abstraction)   & \checkmark & \checkmark & -- & -- \\
$\iota$ (Invariance)     & \checkmark\textsuperscript{\dag} & \checkmark & -- & -- \\
\bottomrule
\end{tabular}
\caption{Operator--principle correspondence. 
\textsuperscript{\ddag}Operators included in the executable conformance suite (\S\ref{app:validation}). 
\textsuperscript{\S}$\rho$ only adds context metadata and does not modify weights, 
so entropy preservation holds by construction; confirmed across all conformance-suite
runs (\S\ref{app:validation}).
\textsuperscript{*}$\delta$ does not strictly preserve weight ratios (P1),
but for $\lambda\in[0,1]$ is entropy-nondecreasing; increase is strict only for
$\lambda>0$ with nonuniform normalized input weights.
\textsuperscript{**}$\sigma_{\mathrm{state}}$ with fixed absolute $b$ is not
operator-scale-equivariant, although its admissible within-state transformation leaves
normalized entropy unchanged.
\textsuperscript{\dag}$\iota$ satisfies P1 by tagged-disjoint retention of all
records from all contexts; this mark makes no unconditional entropy-ordering claim.}
\label{tab:correspondence}
\end{table}

\subsection{Pipeline Composition}

For implementation, a compact transition chain
$\pi \circ \kappa \circ \delta \circ \tau$ is typically sufficient,
with $\phi$ (main text) providing the input transformation from text to
initial state. Optional metadata functions ($\rho, \alpha, \iota$) can be
inserted where context tracking or cross-context labeling is required.
Algorithm~\ref{alg:pipeline} shows this compact pattern with optional
positioning. Each declared transition carries records forward as specified; the suite
separately checks record-weight entropy conformance through repeated updates.

\subsection{Extended Executable Conformance Details}
\label{app:validation}

Section~\ref{sec:operator-validation} presents the main conformance result and summary
table. This appendix gives the full methodology, parameter sweep, figures, and
operator-level results. Each operator is evaluated on the data structure appropriate
to its function, with uniform subtraction serving as the negative comparison.

\subsubsection{Methodology}

\textbf{Data}: Three datasets were constructed using the NRR-Phi 
$\phi$ mapping procedure on a \textbf{newly constructed sentence set, 
entirely distinct from the 68-sentence test set in the main text}:

\begin{enumerate}
    \item \textbf{Single states} ($n = 180$): 60 ambiguous sentences 
          (40 epistemic, 20 lexical; English and Japanese) processed 
          by three LLMs (ChatGPT, Gemini, Claude), yielding 180 states. 
          Each state contains 2--4 retained records with confidence weights.
    
    \item \textbf{Contradictory pairs} ($n = 200$): 150 pairs from the 
          same sentence processed by different models (capturing 
          inter-model disagreement; 50 sentences $\times$ $\binom{3}{2} = 3$ 
          model pairs, selected from the 60 sentences by excluding 
          10 with near-identical model outputs), plus 50 pairs from 
          ambiguous words in different contexts (providing paired lexical-record
          sets without a separate semantic-adequacy adjudication).
    
    \item \textbf{Temporal pairs} ($n = 200$): 150 two-turn dialogue 
          pairs (where context evolves across turns), plus 50 
          context-evolution pairs (where the same sentence is 
          reinterpreted under shifted context).
\end{enumerate}

\textbf{Metric}: Violation rate---percentage of states where
$H(\mathcal{S}') < H(\mathcal{S}) - \epsilon$ with $\epsilon = 0.1$ bits. 
This threshold is 0.1 bits of record-weight entropy; it is not a semantic
information-loss percentage.

\textbf{Operator--data matching}: Weight-transforming transitions 
($\delta$, $\sigma_{\mathrm{state}}$, $\tau$) are tested on single states. The integration 
operator ($\kappa$) is tested on contradictory pairs. The persistence 
operator ($\pi$) is tested on temporal pairs. This matching ensures each 
operator is validated on the data structure it is designed to process.

\textbf{Comparison}: For weight-transforming operators, we compare:
\begin{itemize}
    \item \textbf{Uniform-subtraction comparison}:
          ($\delta$ v1: $w'_i = w_i - b$)
    \item \textbf{Non-violating transitions/checks}: Proportional transformation
          ($\delta$ v2: $w'_i = w_i(1-\lambda) + \bar{w}\lambda$; 
          $\sigma_{\mathrm{state}}$: $w'_i = w_i - b \cdot w_i/\max w$)
\end{itemize}

\textbf{Parameter sweep}: $\delta$ v1 was tested 
at three subtraction magnitudes ($b \in \{0.05, 0.10, 0.20\}$), 
$\delta$ v2 at five compression rates ($\lambda \in \{0.1, 0.2, 0.3, 0.4, 0.5\}$), 
and $\sigma_{\mathrm{state}}$ at four bias levels ($b \in \{0.05, 0.10, 0.15, 0.20\}$). 
This yields 2,740 total operator--state measurements.

\subsubsection{Results}

\begin{figure}[h]
\centering
\includegraphics[width=0.95\textwidth]{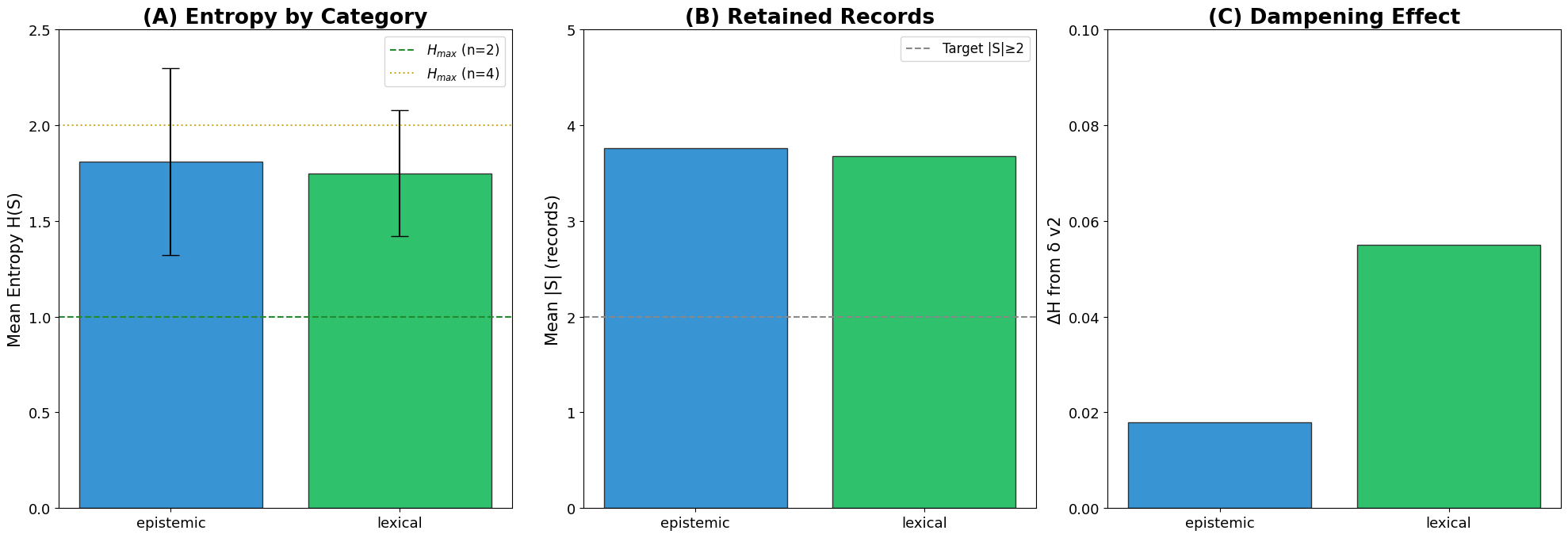}
\caption{Analysis across the reported epistemic and lexical record sets.
(A) Mean entropy by category with standard deviation error bars; both 
categories maintain high multiplicity ($H > 1.7$ bits), approaching the 
theoretical maximum for four positive-weight records ($H_{\max} = 2.0$ bits).
(B) Average number of retained records per state; dashed line
indicates minimum target ($|\mathcal{S}| \geq 2$). 
(C) Entropy increase from dampening operator ($\delta$ v2) relative to 
pre-dampening state, demonstrating consistent preservation across 
both categories.}
\label{fig:categories}
\end{figure}

Figure~\ref{fig:operators} shows entropy change by operator type. 
The principle-violating uniform operator ($\delta$ v1) violates the
record-weight entropy threshold at a rate that increases with subtraction magnitude:
1.7\% at $b = 0.05$, 6.1\% at $b = 0.10$, and 17.8\% at 
$b = 0.20$ (32 of 180 states). In contrast, all tested non-violating
transitions and calibration/identity checks achieve \textbf{0\% violations}
across all parameter settings.

Table~\ref{tab:validation_results} presents the complete results.

\begin{table}[H]
\centering
\begin{tabular}{llrrr}
\toprule
\textbf{Operator} & \textbf{Data} & \textbf{$n$} & \textbf{Violations} & \textbf{Mean $\Delta H$} \\
\midrule
$\delta$ v1 ($b=0.05$) & single & 180 & 3 (1.7\%) & $-0.012$ \\
$\delta$ v1 ($b=0.10$) & single & 180 & 11 (6.1\%) & $-0.030$ \\
$\delta$ v1 ($b=0.20$) & single & 180 & 32 (17.8\%) & $-0.086$ \\
\midrule
$\delta$ v2 ($\lambda=0.1$) & single & 180 & 0 (0\%) & $+0.012$ \\
$\delta$ v2 ($\lambda=0.3$) & single & 180 & 0 (0\%) & $+0.031$ \\
$\delta$ v2 ($\lambda=0.5$) & single & 180 & 0 (0\%) & $+0.044$ \\
\midrule
$\sigma_{\mathrm{state}}$ ($b=0.10$) & single & 180 & 0 (0\%) & $0.000$\textsuperscript{\dag} \\
$\sigma_{\mathrm{state}}$ ($b=0.20$) & single & 180 & 0 (0\%) & $0.000$\textsuperscript{\dag} \\
\midrule
$\tau$ (identity, baseline) & single & 180 & 0 (0\%) & $0.000$ \\
\midrule
$\kappa$ (CPP) & contra.\ pairs & 200 & 0 (0\%) & $+0.881$ \\
$\pi$ (persistence) & temporal pairs & 200 & 0 (0\%) & $+0.922$ \\
\bottomrule
\end{tabular}
\caption{Executable conformance results (representative parameter values shown; 
all omitted settings also achieved 0\% violations). 
The uniform-subtraction comparison ($\delta$ v1) shows increasing violation
rates with subtraction magnitude. All tested non-violating transitions and checks
achieve 0\% violations across all parameter settings tested. $\kappa$ and
$\pi$ show large positive $\Delta H$ because they merge records
from two states, increasing multiplicity. 
\textsuperscript{\dag}$\sigma_{\mathrm{state}}$ applies an identical multiplicative 
factor $({1 - b/\max w})$ to all weights, leaving the normalized 
distribution invariant; $\Delta H = 0$ is analytically exact 
(verified computationally). Similarly, $\tau$ (identity) yields 
$\Delta H = 0$ by mathematical necessity.}
\label{tab:validation_results}
\end{table}

Figure~\ref{fig:categories} presents analysis across the two 
ambiguity categories from the single-state dataset. 
Epistemic ambiguities maintain mean entropy $H = 1.81$ bits 
(SD $= 0.49$) with an average of 3.76 records per state.
Lexical ambiguities show $H = 1.75$ bits (SD $= 0.33$) with 
3.68 records. These structural summaries do not establish that the extracted
records are semantically correct, exhaustive, or mutually distinct readings.

\subsubsection{Key Findings}

\begin{enumerate}
    \item \textbf{Criterion-violation rate scales with subtraction magnitude}: 
          $\delta$ v1 violation rate increases monotonically from 
          1.7\% ($b=0.05$) to 17.8\% ($b=0.20$), confirming that 
          larger uniform subtractions disproportionately suppress
          weaker records.
    
    \item \textbf{Dampening increases entropy in the tested setting}: $\delta$ v2 raises
          $H$ by an average of 0.03 bits at $\lambda=0.3$. In general it is
          entropy-nondecreasing, with strict increase only for $\lambda>0$ and
          nonuniform normalized input weights (Appendix~\ref{app:operators}). The
          reported mean increase is empirical redistribution across extracted records,
          not insertion of unsupported new records.
    
    \item \textbf{State-level calibration is entropy-invariant but not
          operator-scale-equivariant}: 
          $\sigma_{\mathrm{state}}$ yields mean $\Delta H \approx 0$ bits by design,
          because it uniformly rescales weights within each admissible state. With fixed
          absolute $b$, it does not commute with rescaling of the input state.
    
    \item \textbf{Integration and persistence increase record-weight entropy}:
          $\kappa$ and $\pi$ increase $H$ by $+0.88$ and $+0.92$ bits 
          respectively, by merging records from paired states. This is not a
          semantic information-gain claim.
    
    \item \textbf{Zero violations for the tested non-violating transitions and checks}:
          Across 180 single states and 400 pairs, none of the tested non-violating
          transitions or calibration/identity checks exhibited $\Delta H < -0.1$ bits under any parameter
          setting tested.
\end{enumerate}

\section*{Acknowledgments}

The author acknowledges the use of large language models, including Claude (Anthropic), ChatGPT and Codex (OpenAI), and Gemini (Google), for language editing, proofreading, and LaTeX formatting assistance during manuscript preparation. All substantive ideas, claims, analyses, and conclusions are solely the responsibility of the author.

\end{document}